\documentclass[runningheads]{llncs}
\usepackage[T1]{fontenc}
\usepackage{graphicx}
\usepackage{amsmath}
\usepackage{amssymb}
\usepackage{booktabs}
\usepackage{multirow}
\usepackage{microtype}
\usepackage[utf8]{inputenc}
\usepackage{hyperref}
\usepackage{float}
\usepackage{placeins}
\newcommand{\Cskel}{\mathcal{C}_\text{skel}}
\newcommand{\Ctopo}{\mathcal{C}_\text{topo}}
\newcommand{\Cv}{\mathcal{C}_v}
\newcommand{\Cc}{\mathcal{C}_\text{c}}

\begin{document}
\title{Same Branches, Different Trees: A Bifurcation Connectedness Metric for Coronary Artery Segmentation and FFR-CT Decision Agreement}
\titlerunning{Same Branches, Different Trees}
%
\author{Maame Owusu-Ansah\inst{1}\orcidID{0000-0002-2013-5944} \and
Kelvin Lee\inst{2}\orcidID{0000-0001-7705-4486} \and
Vinod Venugopal\inst{2}\orcidID{0009-0000-9683-834X} \and
Moazzam Jawaid\inst{3}\orcidID{0000-0002-8817-0754} \and
Wenting Duan\inst{1}\orcidID{0000-0001-9665-4890} \and
James Brown\inst{1}\orcidID{0000-0001-7636-4554}}
\authorrunning{M. Owusu-Ansah et al.}
\institute{Applied Vision \& Artificial Intelligence Lab, School of Engineering \& Physical Sciences, University of Lincoln, Brayford Pool, Lincoln LN6 7TS, United Kingdom\\
\email{JamesBrown@lincoln.ac.uk} \and
Lincolnshire Heart Centre, United Lincolnshire Teaching Hospitals Trust, Lincoln, United Kingdom \and
School of Computer Science \& Informatics, De Montfort University, The Gateway, Leicester LE1 9BH, United Kingdom}
\maketitle              

\begin{abstract}
Fractional flow reserve derived from CT angiography (FFR-CT) simulates flow through a patient-specific vessel model, so its accuracy depends on the connectedness of the segmented tree, not only on volumetric overlap: a segmentation can reach high Dice yet sever a bifurcation, dropping the downstream subtree and reversing the treatment decision. Topology-aware losses such as clDice and Skeleton Recall act on the global centreline and can miss localised breaks. We study the Bifurcation Connectedness Score (BCS), which scores connectedness at each ground-truth bifurcation, and soft-BCS, its differentiable training surrogate. BCS captures a property of segmentation quality the standard metrics miss: it responds strongly to breaks in connectedness while staying largely unchanged under connectedness-preserving narrowing. Higher BCS accompanies closer agreement between the FFR-CT decisions a solver makes on predicted versus ground-truth geometry, most clearly in severe disease (OR $2.16$, CI $[1.23, 4.18]$). Both decisions come from the same solver, so this reflects geometric, not clinical, fidelity. In training, soft-BCS and Skeleton Recall recover the same branches but build different trees. Recovering branches and keeping them connected are separable properties, so we recommend reporting a measure of each.

\end{abstract}

\section{Introduction}
Coronary artery disease, the narrowing of the heart's arteries by atherosclerotic plaque, is among the leading causes of death worldwide~\cite{roth_global_2020}, and the central question in managing it is whether a narrowing restricts blood flow enough to justify intervention. Fractional flow reserve (FFR), the pressure ratio across a lesion measured during catheterisation, is the reference standard but is invasive and costly~\cite{tonino_fractional_2009}. FFR derived from coronary CT angiography (FFR-CT) estimates the same quantity non-invasively by simulating flow through a patient-specific vessel model, and has been validated against invasive FFR~\cite{taylor_computational_2013,min_diagnostic_2012}.

\begin{figure}[tbp]
\centering
\includegraphics[width=.7\textwidth]{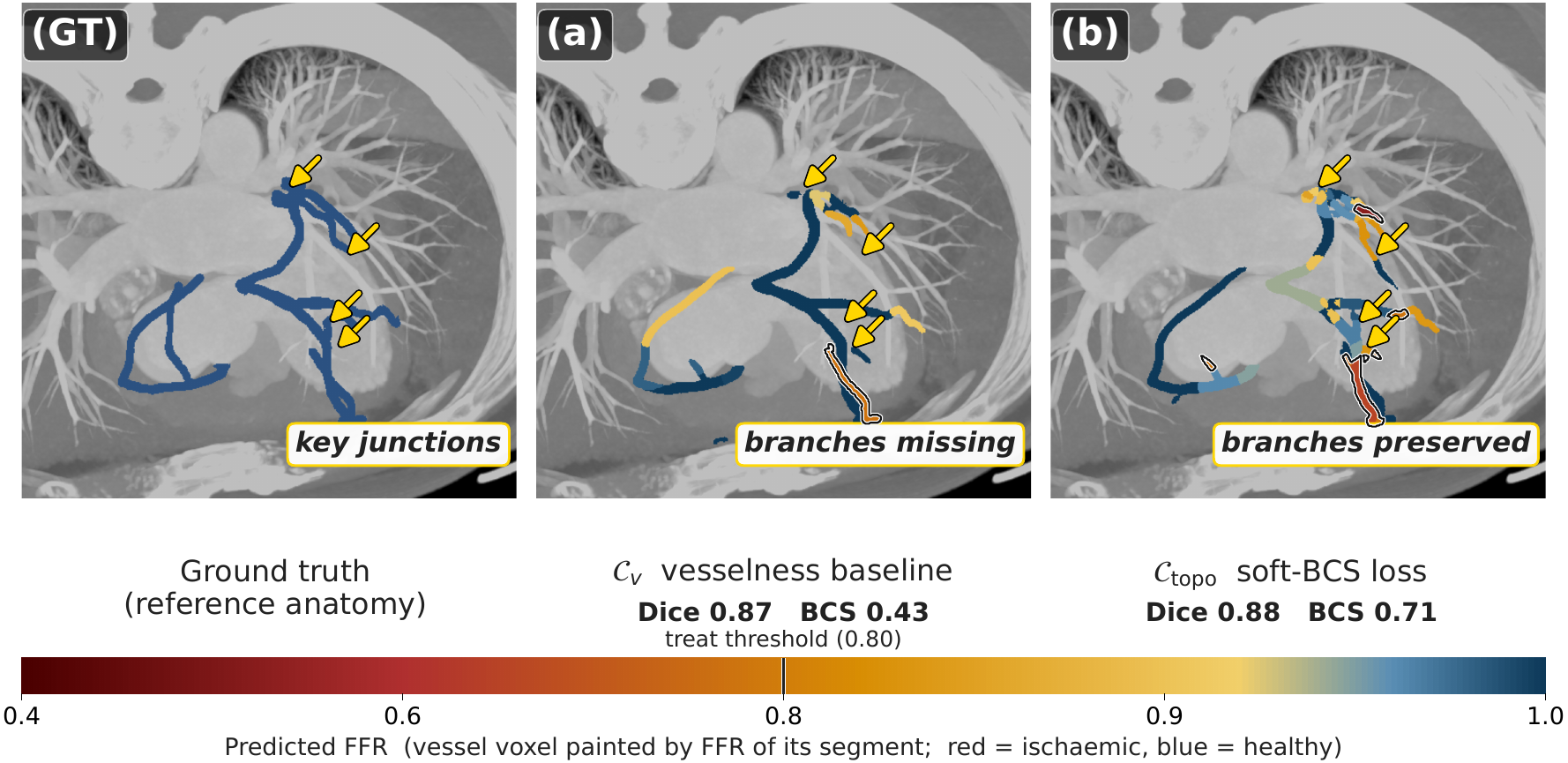}
\caption{Two segmentations with near-identical Dice scores, but vastly different BCS values. \textbf{(a)}~Vesselness baseline ($\Cv$): Dice $0.87$, BCS $0.43$. \textbf{(b)}~Soft-BCS model ($\Ctopo$): Dice $0.88$, BCS $0.71$. Voxels coloured by FFR; red marks ischaemic segments (${\le}0.80$)~\cite{tonino_fractional_2009}. In (a) a disconnected junction drops the downstream branch from the solver, deferring the case; in (b) the junction is preserved and the case is treated. Despite near-identical Dice, the two segmentations differ sharply in BCS.}
\label{fig:def}
\end{figure}

Because FFR-CT simulates flow through the reconstructed tree, its accuracy depends on the connectedness of that tree, not only on how well the segmentation overlaps the true anatomy, and this connectedness is most easily broken at bifurcations. Branch points are both clinically risky and hard to segment: lesions there are harder to treat and carry worse outcomes than those in straight segments~\cite{burzotta_percutaneous_2021}, while the points themselves are small, spanning only $1$--$2$\,mm and a few voxels at typical CT resolution~\cite{lin_deep_2022}, so a minor segmentation error can disconnect a branch entirely. An FFR-CT simulator can then treat that territory as absent, removing its resistance from the flow model and changing the computed FFR (Fig.~\ref{fig:def}). Dice and HD95 measure global overlap, so a single severed junction barely registers against the correctly labelled vessel volume~\cite{maier-hein_metrics_2024}.

The Bifurcation Connectedness Score (BCS) measures whether the branches at each ground-truth bifurcation remain mutually reachable in a prediction. BCS and its surrogate soft-BCS were introduced in preliminary form~\cite{owusu-ansah_adapt_nodate}, which established the metric and loss on a single backbone. Here we characterise BCS across three pretrained architectures, relate it to FFR-CT decision agreement, and show that branch recovery and graph fragmentation dissociate. We also ask what topology-aware losses produce in training: we train soft-BCS and Skeleton Recall alongside clDice and a vesselness baseline and separate which branches a model recovers (Branch Correspondence Ratio, BCR) from how it assembles them ($\beta_0$, the number of disconnected components).

Topology-aware losses can be broadly divided into two distinct categories. clDice~\cite{shit_cldice_2021} is overlap-based, rewarding both recall and precision of skeleton voxels. Skeleton Recall~\cite{kirchhoff_skeleton_2024} and soft-BCS~\cite{owusu-ansah_adapt_nodate} are recall-based: they reward recovering the target branches but do not penalise spurious or disconnected structure, so they control which branches appear but not how they join up. Persistent-homology losses constrain Betti numbers, the counts of connected components ($\beta_0$) and loops ($\beta_1$), in the segmentation, so they regularise overall topology but not connectedness at a specific junction ~\cite{hu_topology-preserving_2019}. Topological errors persist even when Dice is high~\cite{yang_benchmarking_2025}, and coronary methods benchmark topological proxies~\cite{qiu_corsegrec_2023}, but to our knowledge no junction-level metric has been tested against FFR-CT reliability. The three pretrained backbones we evaluate (CT-FM~\cite{pai_vision_2025}, STU-Net~\cite{huang_stu-net_2023}, and SwinUNETR~\cite{tang_self-supervised_2022}) have uncharacterised junction-level connectedness on coronary anatomy.

\subsection*{Contributions}
(i)~BCS measures bifurcation connectedness, a property the standard metrics miss.
(ii)~BCS tracks FFR-CT decision agreement with the reference geometry, sharpest in severe disease; both decisions use the same solver, so this is geometric, not clinical, fidelity.
(iii)~Branch recovery and connectedness are separable: soft-BCS and Skeleton Recall recover equivalent branches but build differently connected trees. We suggest reporting both.

\subsection{Data, Architectures, and Losses}\label{sec:matrix}
We use the ImageCAS dataset~\cite{zeng_imagecas_2023}, consisting of $1{,}000$ coronary CTA volumes split $750/250$ into training and test; all metrics are reported on the $250$ test cases. Volumes are resampled to $1.0$\,mm isotropic spacing and intensities clipped to $[-200, 600]$\,HU. We use three pretrained 3D backbones spanning distinct architectural families, each fine-tuned for coronary artery segmentation: CT-FM~\cite{pai_vision_2025} (SegResNet foundation model, 77M parameters, pretrained on 148{,}000 CT volumes), STU-Net-L~\cite{huang_stu-net_2023} (scaled nnU-Net, 440M, TotalSegmentator), and SwinUNETR~\cite{tang_self-supervised_2022} (Swin-Transformer, 65M, 5{,}050 CT volumes). Every backbone is trained under every loss with three seeds; we report the mean over seeds.

Every configuration optimises the same objective:
\begin{equation}\label{eq:objective}
\mathcal{L} = \lambda_d \mathcal{L}_{\text{Dice}}
            + \lambda_{\text{CE}} \mathcal{L}_{\text{CE}}
            + \lambda_v \mathcal{L}_{\text{Frangi}}
            + \alpha \mathcal{L}_{\text{topo}}.
\end{equation}
The baseline $\Cv$ trains from the pretrained backbone with $\lambda_d{=}0.35$, $\lambda_{\text{CE}}{=}0.15$, $\lambda_v{=}0.35$, $\alpha{=}0$, using the Frangi vesselness term~\cite{frangi_multiscale_1998} and no topology term. The three topology configurations fine-tune from the $\Cv$ checkpoint with $\lambda_d{=}\lambda_{\text{CE}}{=}0.5$, $\lambda_v{=}0$, $\alpha{=}0.05$, differing only in the topology term: clDice~\cite{shit_cldice_2021} ($\Cc$), soft-BCS ($\Ctopo$; Eq.~\ref{eq:softbcs}), or Skeleton Recall~\cite{kirchhoff_skeleton_2024} ($\Cskel$). We use the same topology weight $\alpha{=}0.05$ for all three topology losses, so none is given an advantage by tuning.

All topology configurations share the same $\Cv$ initialisation, data, and $100$-epoch budget. Learning rate is fixed within a backbone ($10^{-4}$ for $\Cv$; $10^{-5}$ for CT-FM, $10^{-4}$ for STU-Net-L and SwinUNETR at the topology stage). We train with AdamW (weight decay $10^{-5}$) on $96^3$ patches with mixed precision and early stopping, sliding-window inference at $0.75$ overlap, and no connected-component post-processing. Runs take $22$--$145$\,h on a single NVIDIA RTX A6000 (48\,GB).

\begin{figure}[t]
\centering
\includegraphics[width=.65\textwidth]{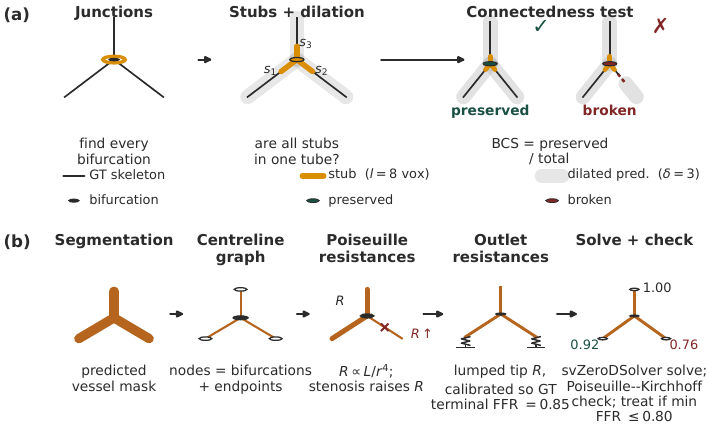}
\caption{Methods overview. (a)~BCS construction: ground-truth junctions are detected, $l{=}8$-voxel stubs are taken along each outgoing branch and tested through the $\delta{=}3$-voxel-dilated prediction; a junction is preserved only if all stubs remain mutually reachable, and BCS is the fraction preserved. (b)~FFR-CT readout: the segmentation is reduced to a centreline graph, each segment is assigned a Poiseuille resistance, lumped outlet resistances are calibrated to a healthy terminal FFR of $0.85$, and an svZeroD solve (Poiseuille--Kirchhoff check) yields per-node FFR with a treat threshold of $0.80$.}
\label{fig:methods}
\end{figure}

\subsection{Bifurcation Connectedness Score}\label{sec:bcs}
BCS~\cite{owusu-ansah_adapt_nodate} is the fraction of ground-truth bifurcations whose branches stay connected in a prediction (Fig.~\ref{fig:methods}a). We skeletonise the ground-truth mask and mark every skeleton voxel with at least three of its $26$ neighbours as a junction, merging touching junction voxels. From each junction we take the first $l{=}8$ voxels of each outgoing branch as a \emph{stub} and dilate the predicted skeleton by $\delta{=}3$ voxels, so a small offset does not count as a break. A junction is preserved when all its stubs remain mutually reachable through the dilated skeleton, and BCS is the fraction preserved. The metric is robust to its two hyperparameters: across $10$ combinations of dilation radius $\delta$ and stub length $l$, the loss ordering by BCS is unchanged.

\subsection{Branch Correspondence and Component Count}\label{sec:bcr}
We report \emph{which} branches a prediction recovers separately from \emph{how connected} they are. The Branch Correspondence Ratio (BCR) measures recovery: we skeletonise both masks, split each skeleton at its bifurcations into branches (dropping those under $4$ voxels as noise), and match ground-truth to predicted branches one-to-one, smallest-distance-first, where two branches match if their symmetric chamfer distance ~\cite{barrow_parametric_1977} is at most $d_{\text{match}}{=}3$\,mm. BCR is the harmonic mean of $\text{cov}_\text{gt}$ and $\text{cov}_\text{pred}$; the fractions of ground-truth and predicted branches matched (recall and precision):
\begin{equation}\label{eq:bcr}
\text{BCR} = \frac{2\,\text{cov}_\text{gt}\,\text{cov}_\text{pred}}
                  {\text{cov}_\text{gt} + \text{cov}_\text{pred}}.
\end{equation}
We summarise connectedness by $\beta_0$, the number of $26$-connected components: an intact tree has $\beta_0{=}1$, with each loose piece adding one. Two predictions can share a BCR yet differ in $\beta_0$, recovering the same branches but assembling them into more or fewer pieces (Sec.~\ref{sec:dissoc}). Branch Endpoint Reachability (BER) is the fraction of ground-truth skeleton endpoints in the largest connected component of the $3$-voxel-dilated predicted skeleton.

\subsection{Soft-BCS Loss}\label{sec:softbcs}
Soft-BCS~\cite{owusu-ansah_adapt_nodate} is a differentiable surrogate for BCS. At each ground-truth junction it rewards high predicted foreground probability along every incident stub but leans on the least-covered one, since a junction is preserved only if \emph{all} its branches are present. For a junction with stubs indexed by $k$,
\begin{equation}\label{eq:softbcs}
\ell = 1 - \sum_k w_k \bar{p}_k, \qquad
w_k = \frac{\exp(-\bar{p}_k / T)}{\sum_j \exp(-\bar{p}_j / T)},
\end{equation}
where $\bar{p}_k$ is the mean foreground probability over stub $k$. The weight $w_k$ is larger for lower $\bar{p}_k$, focusing the loss on the weakest branch; temperature $T{=}0.2$ keeps this close to a hard minimum. The total loss averages $\ell$ over all ground-truth junctions.

\subsection{FFR Readout and Evaluation}\label{sec:ffr}
We estimate FFR-CT by reduced-order flow simulation (Fig.~\ref{fig:methods}b): the predicted centreline graph is cut at bifurcations into segments, each assigned a Poiseuille resistance from its length and radius, with lumped outlet resistances calibrated so the ground-truth tree yields a healthy median terminal FFR of $0.85$. We solve with svZeroDSolver~\cite{pfaller_automated_2022}, verified by an independent Poiseuille--Kirchhoff resistor network, and flag a positive (treat) decision when the minimum FFR anywhere in the tree is ${\le}0.80$~\cite{tonino_fractional_2009}. We compare predicted against ground-truth geometry rather than invasive FFR; per-territory analyses assign each myocardial region to its nearest vessel voxel~\cite{leipsic_scct_2014} to match predicted and ground-truth territory-level FFRs, without altering the binary decision. Our primary cohort is ImageCAS ($N{=}250$); ASOCA~\cite{gharleghi_automated_2022} ($N{=}20$, stenosis without disconnection) serves as a deliberate out-of-distribution test of BCS's calibre-blindness.

\subsection{Statistical Analysis}\label{sec:stats}
To test independence from standard metrics, we regress rank-BCS on Dice, clDice, HD95, and $\beta_0$, reporting $1{-}R^2$ and partial rank correlations. To test the FFR-CT link, we (i) stratify cases into BCS quartiles, reporting patient-clustered bootstrap confidence intervals (B{=}2{,}000, resampled by case) rather than row-level p-values, and (ii) fit a patient-clustered GEE ($N{=}250$) entering all six metrics (BCS, Dice, clDice, HD95, $\beta_0$, BER), reporting odds ratios (OR) per standard deviation for decision error so that BCS is adjusted for the other five; we read these as adjusted associations within a correlated metric stack, not marginal effects. Loss comparisons use Wilcoxon signed-rank (continuous) and McNemar's test (decisions); we correct for multiple comparisons with Holm's method ($p{<}0.05$) and report $d_z$ as paired effect size.

\section{Results}
\subsection{BCS measures something the standard metrics do not}\label{sec:distinct}
BCS measures a property of a coronary segmentation that the standard metrics cannot see: whether the vessel tree actually stays connected at its bifurcations. Statistically, BCS is largely independent of them. Regressing rank-BCS on Dice, clDice, HD95, and $\beta_0$ leaves $78.7\%$ of its variance unexplained ($R^2{=}0.21$); pairwise, it correlates moderately with Dice ($r{=}0.45$) and not at all with $\beta_0$ ($r{=}-0.07$), and every partial correlation stays below $0.30$. \textbf{Perturbation specificity.} Severing a branch drops BCS $8.7{\times}$ more than Dice; connectedness-preserving narrowing leaves it unchanged ($|\Delta\text{BCS}|{=}0.004$; $N{=}750$, all three architectures). BCS separates a break from a narrowing ${\sim}20{\times}$, against ${\le}5{\times}$ for clDice, Dice, and HD95; $\beta_0$ peaks on pruning rather than severing.

\textbf{BCS catches failures the other metrics miss.} In $93$ patient-architecture-seed triples, Dice is high (${\ge}0.80$) yet BCS is low (${<}0.50$). One might expect $\beta_0$ to catch these breaks, but it does not: in this subgroup $\beta_0$ is actually \emph{lower} than the test-set average ($4.01$ vs.\ $4.56$). BCS alone separates the subgroup from the rest (mean BCS $0.437$ vs.\ $0.785$). The pattern holds across all three backbones.

\subsection{BCS and FFR-CT decision agreement}\label{sec:clinical}
Segmentations that better preserve bifurcation connectedness yield FFR-CT decisions closer to those computed on the true geometry. Because both decisions come from the same solver, this reflects geometric fidelity to the reference, not clinical FFR accuracy. The agreement is strongest where connectedness matters most: in severe disease ($\text{FFR}_{\text{GT}}{<}0.75$), consistency rises from low- to high-BCS quartiles ($80.6\%$ to $90.0\%$; OR $2.16$, patient-clustered CI $[1.23, 4.18]$). It weakens in borderline cases (OR $1.47$, CI $[0.52, 4.07]$): BCS stratifies risk across a cohort but is not a per-case verdict.

Across the full cohort the association is weaker and uneven across quartiles (Table~\ref{tab:clinical}). Because the metrics are strongly correlated, no single one can be cleanly credited; we therefore fit a patient-clustered model (a generalised estimating equation, GEE) that adjusts each metric for the others. Only BCS predicts higher decision agreement after adjustment (OR $0.81$ per SD, $p{=}0.002$); Dice shows the same trend without significance (OR $0.81$, $p{=}0.12$), and clDice is uninterpretable under the collinearity. We draw no ranking from these coefficients, only that BCS survives adjustment.

ASOCA tests the opposite failure mode: stenosis without disconnection. BCS should stay flat while overlap metrics suffer, and it does: across $N{=}20$ cases Dice falls $31$--$39$\,pp while BCS rises $+2.8$ to $+10.5$\,pp. The cohort is too small to test the FFR link, so we limit the FFR claim to anatomy where vessel-tree structure, not calibre, drives the decision.

\begin{table}[t]
\centering
\caption{BCS and FFR-CT decision consistency, pooled across architectures, losses, and seeds. Left: consistency by BCS quartile ($95\%$ Wilson intervals); not monotonic (Q2 below Q1). Right: multivariable GEE odds ratios for decision \emph{error} per $+1$\,SD, all six metrics and patient clustering. BCS is the only metric with a significant protective association under clustering (OR~$0.81$, $p{=}0.002$); Dice shares the point estimate; clDice OR~${>}1$ reflects collinearity.}
\label{tab:clinical}
\setlength{\tabcolsep}{6pt}\small
\begin{tabular}{lcc@{\hskip 18pt}lcc}
\toprule
BCS quartile & $n$ & Consistency & Predictor & OR/SD & $p$ \\
\cmidrule(lr){1-3}\cmidrule(lr){4-6}
Q1 (low)  & 937 & $77.2\%$ & BCS    & $\mathbf{0.81}$ & $\mathbf{0.002}$ \\
Q2        & 944 & $74.7\%$ & BER    & $0.86$ & $0.02$ \\
Q3        & 978 & $83.1\%$ & clDice & $1.41$ & $0.03$ \\
Q4 (high) & 889 & $85.0\%$ & Dice   & $0.81$ & $0.12$ \\
\cmidrule(lr){1-3}
Q4 vs Q1 & \multicolumn{2}{c}{OR $1.68$} & HD95 / $\beta_0$ & n.s. & \\
\bottomrule
\end{tabular}
\end{table}

\subsection{Recall and organisation dissociate}\label{sec:dissoc}
\begin{figure}[tbp]
\centering
\includegraphics[width=.5\textwidth]{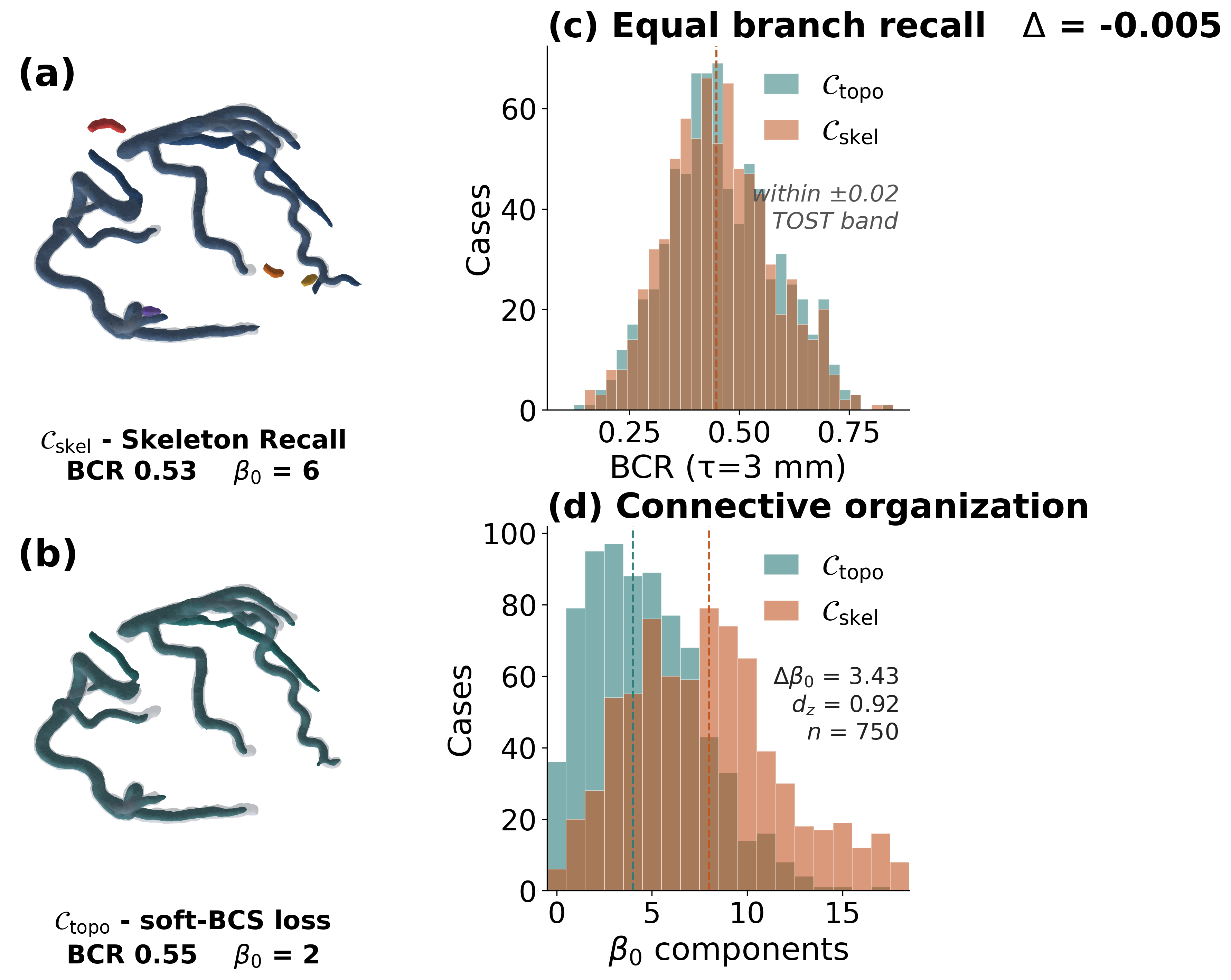}
\caption{Same branches, different trees. (a,~b) For one representative case, $\Cskel$ and $\Ctopo$ recover the same ground-truth branches (equal BCR) but $\Cskel$ scatters them into more disconnected components while $\Ctopo$ keeps a single tree. (c,~d) Paired distributions: BCR is equivalent across losses, while $\beta_0$ is higher for $\Cskel$.}
\label{fig:dissociation}
\end{figure}

Skeleton Recall ($\Cskel$) and soft-BCS ($\Ctopo$) recover the same branches but connect them differently (Fig.~\ref{fig:dissociation}; Table~\ref{tab:main}). BCR differs by $-0.0037$ (95\% bootstrap CI $[-0.009,{+}0.002]$), within a $\pm0.02$ margin corresponding to the maximum BCR seed-level standard deviation (${\le}0.018$, Table~\ref{tab:main}); differences at this scale are within training noise. Connectedness is not equivalent: pooled over $2{,}250$ patient-architecture-seed triples, $\Cskel$ leaves more disconnected components than $\Ctopo$ ($\Delta\beta_0{=}+2.94$, $d_z{=}0.68$, $p{<}10^{-10}$) in $70.8\%$ of cases, consistently across all architectures. The FFR solver bridges gaps below $1.5$\,mm before computing flow, so this fragmentation does not fully propagate to the treat-or-defer decision. At equal BCR and near-equal FFR-CT agreement, $\Cskel$ and $\Ctopo$ look interchangeable on recall metrics, yet only $\beta_0$ reveals that $\Ctopo$ reaches that recall with a far more intact tree.

\subsection{Comparing topology-aware losses}\label{sec:bench}
Table~\ref{tab:main} reports all twelve configurations. $\Cc$ leaves FFR-CT consistency at or below the no-topology baseline on every architecture; both recall-based losses raise it ${\sim}10$\,pp (pooled $p{=}8{\times}10^{-11}$), agreeing to within a point on full-seed architectures. This gain comes at some cost to boundary accuracy ($\Cskel$ carries the highest HD95 and $\beta_0$; Table~\ref{tab:main}), while Dice stays within $0.011$ of baseline for every loss: no single loss dominates all axes.

\begin{table}[tbp]
\scriptsize
\setlength{\tabcolsep}{3pt}
\renewcommand{\arraystretch}{0.85}
\centering
\caption{ImageCAS test set ($N{=}250$), mean over $3$ seeds. $\Cv$: no-topology baseline; $\Cc/\Ctopo/\Cskel$: clDice/soft-BCS/Skeleton Recall. \textbf{Bold}: best per column within each architecture. Seed std: ${\le}0.018$ (rates), ${\le}2.0$ (HD95), ${\le}1.1$ ($\beta_0$), ${\le}4.5$\,pp (FFR).}
\label{tab:main}
\setlength{\tabcolsep}{3pt}
\footnotesize
\begin{tabular}{ll ccc ccc c}
\toprule
 &  & \multicolumn{3}{c}{Overlap / boundary} & \multicolumn{3}{c}{connectedness} & Clinical \\
\cmidrule(lr){3-5}\cmidrule(lr){6-8}\cmidrule(lr){9-9}
Arch. & Loss & Dice\,$\uparrow$ & HD95\,$\downarrow$ & clDice\,$\uparrow$ & BCS\,$\uparrow$ & BCR\,$\uparrow$ & $\beta_0$\,$\downarrow$ & FFR\,$\uparrow$ \\
\midrule
\multirow{4}{*}{CT-FM}
 & $\Cv$     & \textbf{0.809} & 6.00 & 0.875 & 0.728 & 0.467 & 4.3 & 75.4 \\
 & $\Cc$     & 0.809 & 5.91 & \textbf{0.878} & 0.728 & \textbf{0.470} & \textbf{4.2} & 75.3 \\
 & $\Ctopo$  & 0.807 & \textbf{5.63} & 0.872 & 0.757 & 0.460 & 4.6 & 85.5 \\
 & $\Cskel$  & 0.808 & 6.44 & 0.861 & \textbf{0.794} & 0.455 & 8.0 & \textbf{85.7} \\
\midrule
\multirow{4}{*}{STU-Net-L}
 & $\Cv$     & 0.818 & \textbf{5.19} & \textbf{0.890} & 0.758 & \textbf{0.499} & 2.9 & 77.5 \\
 & $\Cc$     & 0.814 & 5.49 & 0.888 & 0.750 & 0.496 & \textbf{2.6} & 76.6 \\
 & $\Ctopo$  & 0.813 & 7.03 & 0.879 & 0.765 & 0.476 & 4.0 & \textbf{86.8} \\
 & $\Cskel$  & \textbf{0.820} & 5.71 & 0.874 & \textbf{0.815} & 0.480 & 5.8 & 86.7 \\
\midrule
\multirow{4}{*}{SwinUNETR}
 & $\Cv$     & \textbf{0.799} & 6.02 & 0.874 & 0.744 & 0.466 & 7.9 & 78.5 \\
 & $\Cc$     & 0.797 & \textbf{5.99} & \textbf{0.878} & 0.738 & \textbf{0.472} & \textbf{5.6} & 75.4 \\
 & $\Ctopo$  & 0.788 & 8.95 & 0.859 & 0.742 & 0.435 & 8.3 & \textbf{88.0} \\
 & $\Cskel$  & 0.798 & 11.52 & 0.850 & \textbf{0.780} & 0.425 & 11.9 & 85.1 \\
\bottomrule
\end{tabular}
\end{table}
\FloatBarrier
\section{Discussion and Conclusion}
Standard metrics measure volumetric overlap, whereas BCS measures bifurcation connectedness ($R^2{=}0.21$, Sec.~\ref{sec:distinct}). BCS stratifies decision consistency most sharply in severe stenosis, where connectedness errors most directly alter the simulated flow path; we note this reflects geometric fidelity to the reference, not clinical FFR accuracy (see Limitations). We therefore argue that the contribution of soft-BCS is not that it outperforms $\Cskel$ on raw BCS (Table~\ref{tab:main} shows $\Cskel$ scores higher), but that the two recall-based losses, despite recovering equivalent branches, dissociate in how they organise them ($\Delta\beta_0{=}+2.94$). Because branch recovery and connectedness can diverge, we recommend reporting BCR, $\beta_0$, and BCS together.

\textbf{Limitations.} BCS is insensitive to calibre-only errors and should be read alongside overlap metrics. The FFR analysis measures geometric, not physiological, fidelity under a simplified resistor model; validation against invasive FFR is needed. A promising next step is to test whether BCS-flagged connectivity failures coincide with clinical FFR-CT/invasive-FFR disagreements: if severed bifurcations drive a subset of those discordant cases, BCS would offer a pre-computation flag for them.

\bibliographystyle{splncs04}
\bibliography{references}
\end{document}